\documentclass{bmvc2k}
\usepackage[small]{caption}
\usepackage{booktabs}       

\usepackage{mathtools}
\usepackage{xspace}

\makeatletter
\DeclareRobustCommand\onedot{\futurelet\@let@token\@onedot}
\def\@onedot{\ifx\@let@token.\else.\null\fi\xspace}
\def\etc{etc\onedot} 
 
\makeatother

\title{Multi-Task Deep Networks for Depth-Based 6D Object Pose and Joint Registration in Crowd Scenarios}

\addauthor{Juil Sock}{ju-il.sock08@imperial.ac.uk}{1}
\addauthor{Kwang In Kim}{k.kim@bath.ac.uk}{2}
\addauthor{Caner Sahin}{c.sahin14@imperial.ac.uk}{1}
\addauthor{Tae-Kyun Kim}{tk.kim@imperial.ac.uk}{1}

\addinstitution{
 Imperial Computer Vision and Learning Lab (ICVL),
 Imperial College,
 London,
 UK
}
\addinstitution{
 Department of Computer Science,
 University of Bath,
 Bath,
 UK
}

\runninghead{Sock et al.}{6D Pose Estimation and Joint Registration in Crowd Scenarios}

\def\eg{\emph{e.g}\bmvaOneDot}

\def\etal{\emph{et al}\bmvaOneDot}

\begin{document}

\maketitle

\begin{abstract}
In bin-picking scenarios, multiple instances of an object of interest are stacked in a pile \textit{randomly}, and hence, the instances are inherently subjected to the challenges: \textit{severe occlusion}, \textit{clutter}, and \textit{similar-looking distractors}.
Most existing methods are, however, for single isolated object instances, while some recent methods tackle crowd scenarios as post-refinement which accounts multiple object relations. 
In this paper, we address recovering 6D poses of multiple instances in bin-picking scenarios in depth modality by multi-task learning in deep neural networks. Our architecture {\it jointly} learns multiple sub-tasks: 2D detection, depth, and 3D pose estimation of individual objects; and joint registration of multiple objects. For training data generation, depth images of physically plausible object pose configurations are generated by a 3D object model in a physics simulation, which yields diverse occlusion patterns to learn. We adopt a state-of-the-art object detector, and 2D offsets are further estimated via a network to refine misaligned 2D detections. The depth and 3D pose estimator is designed to generate multiple hypotheses per detection. This allows the joint registration network to learn occlusion patterns and remove physically implausible pose hypotheses. We apply our architecture on both synthetic (our own and Sil\'eane dataset) and real (a public Bin-Picking dataset) data, showing that it significantly outperforms 
state-of-the-art methods by 15-31 \% in average precision.
\\

\vspace{-1.em}
\end{abstract}
\vspace{-1.5em}
\section{Introduction}
\vspace{-0.5em}
\label{sec:intro}
6D object pose recovery has extensively been studied in the past decade as it is of great importance to many real world applications, such as scene interpretation, augmented reality, and particularly, robotic grasping and manipulation. Robotic manipulators can facilitate industry-related processes, \eg, warehousing, material handling, packaging, picking items from conveyors, shelves, pallets, and placing those to desired locations, as demonstrated in Amazon Picking Challenge (APC) \cite{eppner2016lessons}.\\
\indent A special scenario for robotic manipulation is bin-picking where multiple instances of an object of interest exist in the scene. Being randomly stacked in a box, or on a pallet, the instances are inherently subjected to severe occlusion and clutter, thus making the estimation of the 6D poses of multiple instances challenging.\\
\indent Most existing methods are for single object instance~\cite{posechallenge2017}, while few recent methods tackle crowd scenarios in post-refinement which accounts multiple object relations~\cite{doumanoglou2016recovering}. Prior-arts for 6D object pose include template-based methods~\cite{hinterstoisser2011linemod}, point-pair features~\cite{drost2010model}, or per-pixel regression/patch-based on top of Hough forests~\cite{tejani2014latent,doumanoglou2016recovering,brachmann2014learning,krull2015learning}. These methods are mostly applied and tested on single object scenario with mild occlusion. More recent methods adopt CNNs for 6D pose estimation, taking RGB images as inputs. BB8 \cite{Rad_2017_ICCV} and Tekin~\etal~\cite{tekin2017real} perform corner-point regression followed by PnP for 6D pose estimation. SSD6D \cite{Kehl_2017_ICCV} uses SSD for 2D bounding box detection, while depth is directly inferred from the scales of bounding boxes, and applies a network with a softmax layer for 3D pose. The method has shown comparable performance to other RGBD-based methods, however, fails in the occlusion dataset \cite{brachmann2014learning}. Typically employed is a computationally expensive post processing step such as iterative closest point (ICP) or a verification network \cite{Kehl_2017_ICCV,Rad_2017_ICCV}. 

For the crowd scenarios, \textit{i.e.}, bin-picking, rather than considering each hypothesis individually, \cite{aldoma2012global} formulates a joint registration problem. It obtains an optimal set of hypotheses of multiple objects by minimizing the cost function. Doumanoglou~\etal~\cite{doumanoglou2016recovering} collect a real dataset of bin-picking scenarios, proposing 6D pose estimator based on a sparse auto-encoder and hough forest, and then tackling the joint registration in the similar way \cite{aldoma2012global}. Recently, \cite{bregier2017iccv} proposes datasets of both synthetic and real in bin-picking scenarios, and \cite{Park_2017_ICCV_Workshops} tackles deformable objects (fruits) in crowd by matching local descriptors.\\
\indent 
In the domain of 2D object detection, recent methods have shown that learning relation between nearby hypotheses in crowd scenario can increase the detection performance by effectively suppressing false hypotheses. \cite{Hosang_2017_CVPR} proposes a method to learn non maximum suppression for 2D bounding boxes. More recently, \cite{Hu2018relation_network} trains a 2D detection network jointly with relation module in end-to-end manner.\\

\begin{figure}[t]
\centering
\includegraphics[width=\linewidth]{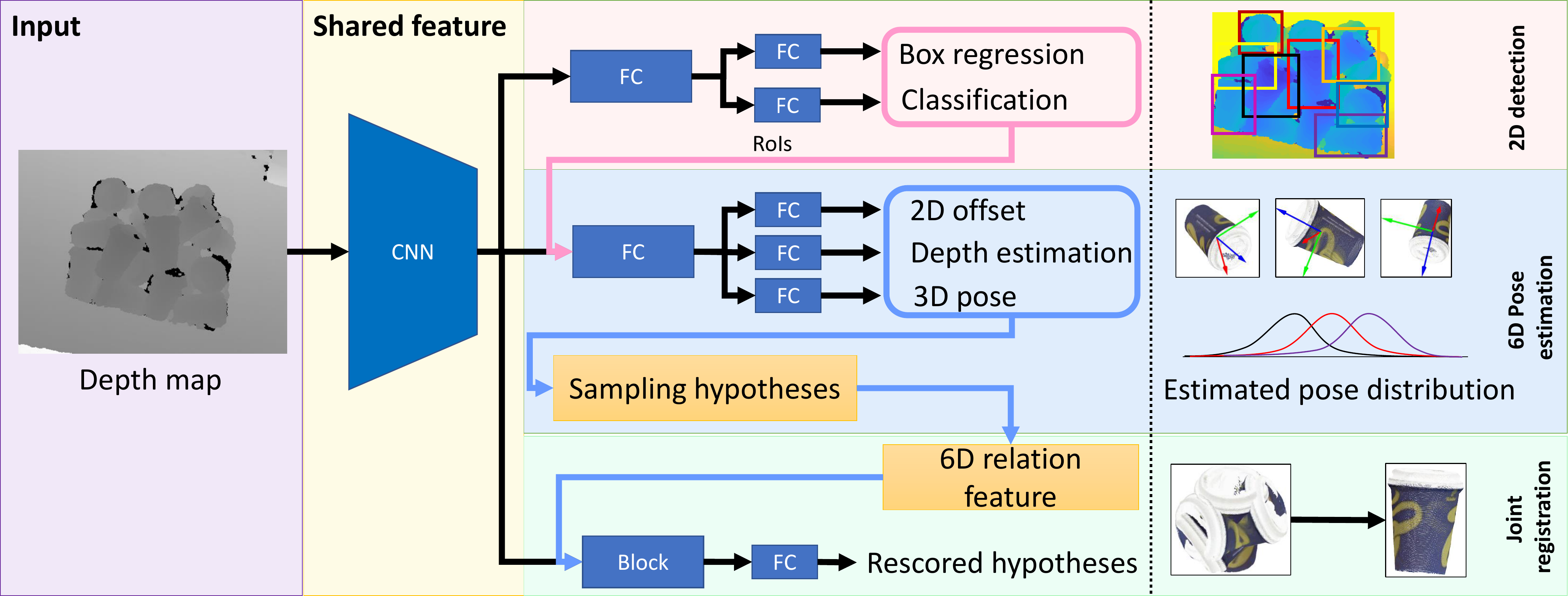}
\vspace{0.2em}
\caption{Our multi-task learning network architecture consisting of three modules each optimized for different sub-tasks: 2D detection, 6D pose estimation, and joint registration. During inference, each sub-tasks are sequentially performed (red and blue arrows). During training, the entire network is trained jointly using multi-task losses, generating the convolutional features shared across sub-tasks.}
\label{fig:overview}
\vspace{-2.0em}
\end{figure}
\indent In this paper, we propose a depth-based 6D pose estimation network which jointly detects and directly recovers (without ICP) full 6D poses from single depth images. Depth image is used in the proposed architecture, as instances are densely populated in crowd scenario and geometric information from depth image is desirable. Directly estimating multiple 6D poses of objects from a single image is a difficult task, therefore the architecture has modules optimized for different sub-tasks: image 2D detection, depth estimation and 3D pose estimation for individual objects, and joint registration of multiple objects as shown in Fig.~\ref{fig:overview}. Although the full architecture consists of several modules, the tasks are jointly learned. The proposed network explicitly learns physically plausible object pose configurations and occlusion patterns from depth images generated by physics simulator and 3D models of objects.

2D offset estimation regresses 2D center position of the object given the detected box coordinate to compensate misalignment during the bounding box regression. Multiple hypotheses per detected object are generated by the depth and 3D pose estimator by formulating the estimation problem as a classification task. Lastly, relations between hypotheses from the pose estimator are modeled by the joint registration network using block modules \cite{Hosang_2017_CVPR} to suppress false positives. Our architecture is evaluated on both synthetic and real datasets demonstrating that it significantly outperforms state-of-the-art methods by 15-31 \% in AP. 
\vspace{-1em}
\section{Related Work}
\label{sec:Literature}
In this section, we first review 6D object pose recovery for \textit{single object instances}, and then for \textit{crowd scenarios}, \textit{i.e.}, bin-picking. Lastly, we present studies on \textit{relational reasoning}, which is used to further improve the confidence of a 6D hypothesis in our study.\\
\noindent \textbf{Single object instances.} 
Under moderate occlusion and clutter, accurate pose estimation has been achieved by approaches with handcrafted features~\cite{hinterstoisser2011linemod,drost2010model,tejani2014latent}. 3D feature proposed in~\cite{drost2010model} can deal with clutter and occlusion but the performance depends on the complexity of the scene and it is computationally expensive. A holistic template matching approach, Linemod \cite{hinterstoisser2011linemod}, exploits color gradients and surface normals to estimate 6D pose. However, it is sensitive to occlusion as it is difficult to match occluded object to template captured from a clean view of the object. More recently, per-pixel regression/patch-based approaches~\cite{tejani2014latent,doumanoglou2016recovering,brachmann2014learning,krull2015learning} has shown robustness across occlusion and clutter. ~\cite{krull2015learning,brachmann2014learning} use random forest to obtain pixel-wise prediction of class and object coordinate followed by RANSAC. However, performance was only evaluated on scenes with mild occlusion and limited variability of object poses relative to the camera. LCHF\cite{tejani2014latent} and Doumanoglou~\etal~\cite{doumanoglou2016recovering} use patch-based approach on top of hough voting and has shown strong performance under moderate occlusion. However, detection of objects under strong occlusion is difficult as the occluded object only receives vote from visible portion in hough space.\\
\indent 
More recent paradigm on recovering 6D object pose, is to formulate the problem using neural networks\cite{Balntas_2017_ICCV}, jointly learning 6D pose estimation in RGB-only images \cite{Kehl_2017_ICCV,tekin2017real,xiang2017posecnn}. \cite{Kehl_2017_ICCV} extends 2D object detector to simultaneously detect and estimate pose and recover 3D translation by precomputing bounding box templates for every discrete rotation. However, predicted bounding boxes can be highly variable in the presence of occlusion which may lead to incorrect 3D translation estimation. Rad~\etal~\cite{Rad_2017_ICCV} regress 3D bounding box coordinates in 2D image plane and utilize PnP algorithm to recover the full 6D pose. However, the method uses segmentation masks to detect objects, and thus making the hypotheses unreliable when objects of the same class are in high proximity. Despite being robust across occlusion and clutter under non-crowd scenarios, the methods \cite{Kehl_2017_ICCV, Rad_2017_ICCV} require a final refinement step in order to generate accurate hypotheses. \cite{tekin2017real} does not need post-processing. However, the method aggregates estimation from multiple cells under assumption that adjacent cells in the tessellated input image belong to the same object, limiting the scalability of the method to isolated scenarios. \\

\noindent \textbf{Crowd scenarios.} 
Bin-picking scenario is a very challenging task for any vision algorithm since it exhibits several cascading issues (\eg, severe occlusions, foreground clutter,~\etc). The method proposed by Liu~\etal~\cite{liu2012fast} extracts robust depth edges and employs fast directional chamfer matching to reliably detect objects and estimate their poses. 
\cite{aldoma2012global} obtains an optimal set of hypotheses of multiple objects by minimizing a cost function, rather than considering each hypothesis individually. \cite{doumanoglou2016recovering} proposes 6D pose estimator of individual objects based on a sparse auto-encoder and hough forest tackling the joint registration in the similar way as \cite{aldoma2012global}.  However, the method computationally demanding with increasing number of hypotheses. 
\cite{Park_2017_ICCV_Workshops} tackles deformable objects (fruits) in the crowd by matching local descriptors. Recently, Zeng~\etal~\cite{zeng2017multi} address the challenges of the crowd scenarios proposing a self-supervised deep learning method for 6D pose estimation in Amazon Picking Challenge.\\
\noindent \textbf{Relational reasoning.} Although most modern object detectors generate detection hypotheses independently, recent works have shown exploiting the relation between hypotheses helps under high occlusion. \cite{lee2016individualness} models relation between detection hypotheses and uses determinantal point process to select the most confident hypothesis. \cite{stewart2016end} proposes a network for person detection which eliminates the non maximum suppression stage by modeling relation between detections using LSTM \cite{hochreiter1997long}.
Gossipnet\cite{Hosang_2017_CVPR} performs non-maximum suppression by proposing a block module which develops a representation for each detection which accounts for other nearby detections. \cite{Hu2018relation_network} adds an extra layer called relation module to the standard object detection framework and jointly trains in an end-to-end fashion. It has shown that learning object detector with relational reasoning improves detection performance. Both methods \cite{Hosang_2017_CVPR,Hu2018relation_network} use relational reasoning to remove duplicate hypotheses in 2D detection. In this work, the architecture developed in \cite{Hosang_2017_CVPR} is extended to 6D pose estimation problem.

\vspace{-1.5em}
\section{Proposed Architecture}
\label{sec:Methods}
\begin{figure}[t]
\centering
\includegraphics[width=\linewidth]{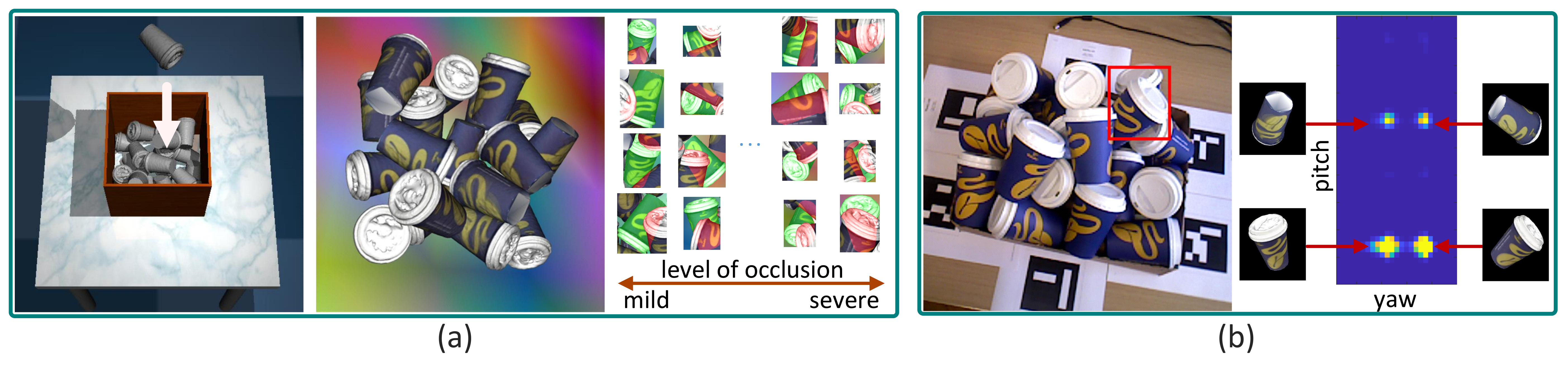}
\vspace{-0.5em}
\caption{\textbf{(a)} Training data generation process: (left) A cluttered scene generated via physical simulation; (middle) An example training view rendered from the scene in (left); (right) Example occlusion patterns (for visualization only; our network is trained on holistic images). \textbf{(b)} (left) An example detection under severe occlusion (overlaid on the RGB image for visualization); (right) Multiple hypotheses with confidences (the brighter is the higher in the heatmap) generated by the 6D pose estimator.}
\label{fig:distribution}
\vspace{-1.50em}
\end{figure}

\vspace{-0.5em}
A key challenge in object pose estimation in bin-packing scenarios is that multiple object instances exhibit severe occlusions under environments cluttered with the same object. For severely occluded objects, the 6D pose predictions made based on visible parts are inherently ambiguous (see~Fig.~\ref{fig:distribution}(b)). Therefore, a joint registration step exploiting the neighborhood context (of how the pose of an object is related to neighboring objects) is essential to making reliable predictions. Existing approaches address this issue only at the testing stage, \eg, using non-maximum suppression (NMS) and the training of their pose estimators is agnostic to such occlusion behaviors: They are trained on isolated object instances. Instead, we directly incorporate the variety of occlusion patterns and the possibility of exploiting the neighborhood context into the \emph{training} phase. To facilitate this process, we construct a new training dataset that reflects such a realistic occlusion-heavy application environment.

Figure~\ref{fig:overview} shows our pose estimator architecture consisting of three modules, each instantiated as a deep neural network: The detection module (red) estimates 2D (along the image plane in the depth map) object locations as bounding boxes. Within each hypothesized bounding box, 6D pose estimation module (blue) estimates the center of the object (which may differ from the center of its bounding box) and its depth (completing the 3D locations), and based on them, generates multiple 3D pose hypotheses (as singling out a correct pose is challenging under heavy occlusions). 
Finally, the joint registration module (green) disambiguate between multiple 6D pose hypotheses and eliminate false positives by considering local neighborhood structures. Exploiting the dependency present in these three tasks and the corresponding modules, we formulate their joint training as multi-task learning (MTL): All three modules share the same depth feature extractor (yellow block in Fig.~\ref{fig:overview}), enabling the training of convolutional features in a unified manner.\footnote{As there are non-differentiable paths joining the three modules (along the red and blue arrows in Fig. \ref{fig:overview}), the training is not end-to-end.}

The overall loss function (of network weights) is defined per depth image as:
\vspace{-0.7em}
\begin{align}
\label{eqn:overall_loss}
\begin{split}
L=L_{D}+L_{O}+L_{D}+L_{P},
\end{split}
\end{align}
with $L_{D}$, $L_{O}$, $L_{D}$, and $L_{P}$ being the losses for estimating 2D bounding boxes, 2D object centers, 3D depths, and 6D poses, respectively. During training, the backward propagation takes place as usual for each module, and they are subsequently combined to update the shared feature extractor in Fig.~\ref{fig:overview}. The 2D detection and 6D pose estimation modules are simultaneously updated while for the update of the joint registration module, we hold the rest of the network weights fixed for improved convergence. The remainder of this section details each module and the corresponding individual loss function in Eq.~\ref{eqn:overall_loss}.
\vspace{-0.8em}
\paragraph{Generating training images.}
Unlike most existing 6D pose estimators which learn object appearance from \emph{isolated views} of objects, our system jointly learns object appearances and their mutual occlusion patterns based on cluttered scene images. Figure~\ref{fig:distribution}(a) illustrates the generation of training data in this scenario: First, we construct a cluttered scene in a virtual environment by randomly stacking the object of interest in a box similarly to~\cite{Sock_2017_ICCV_Workshops,bregier2017iccv}(Fig.~\ref{fig:distribution}(a) left). Each object has 2,000 different bin picking scenes, each of which has 10 to 20 instances in the virtual bin. Then, we render training depth images by sampling virtual camera views (Fig.~\ref{fig:distribution}(a) middle): The camera is always facing towards the center of the bin but with randomized in-plane rotations to capture various occlusion patterns. 17 depth images are generated from each scene constituting a set of 34K images per object.
\vspace{-0.8em}
\paragraph{2D bounding box detection.}
In a bin-picking scenario, the number of objects in a given scene is not known a priori. Therefore, the traditional detection strategy of predicting the single most probable location is not applicable. We employ Faster RCNN~\cite{ren2015faster} which has shown its effectiveness in detecting multiple objects in cluttered scenes. While Faster RCNN was originally developed for RGB images, applying it to depth maps is straightforward. The network receives a single image and generates a bounding box and its confidence per object instance: During training, it minimizes the box regression error (via $L_1$-loss), and for the detection confidence, the soft classification error (via cross entropy) of foreground and background. We denote the combination of the two losses as $L_{D}$ (see Eq.~\ref{eqn:overall_loss}).
\vspace{-0.8em}
\paragraph{6D pose estimation.}
The 6D pose of an object consists of its 3D location and 3D pose (orientation). Given the bounding box of each object instance estimated by 2D detection module, 6D pose estimation module generates multiple hypothesized 6D poses, which are subsequently fed to the joint registration module.

\emph{2D object center estimation.} For each bounding box, we first estimate the accurate 2D location of the corresponding object center. In existing 6D pose estimation approaches (developed for uncluttered scenes)~\cite{Kehl_2017_ICCV, Do2017deep6dpose}, the center of an estimated box is considered identical to the center of the object it contains. For uncluttered scenes, this is reasonable as often, the estimated boxes are well-aligned with the underlying tight ground-truth bounding boxes. However, we experimentally observed that for objects under severe occlusion, such accurate bounding box estimation is extremely challenging, and na\"{i}vely basing the joint registration on such erroneous 2D location estimates eventually leads to erroneous 6D pose predictions (see Table \ref{tab:result}(a) `Ours (w/o offset)').
Therefore, we construct an additional 2D regression network that explicitly estimate the object centers by minimizing the deviation from the ground-truth within each bounding box:
\vspace{-0.5em}
\begin{align}
\small
\label{eqn:parameterised_offset}
L_{O}=\bigg\|\bigg[\frac{x-c_{x}}{w},\frac{y-c_{y}}{h}\bigg]^\top -\bigg[\frac{x^{*}-c_{x}}{w},\frac{y^{*}-c_{y}}{h}\bigg]^\top\bigg\|_1,
\end{align}
where $(x,y)$ and $(x^{*},y^{*})$ denote the predicted and the ground-truth object center coordinates, respectively, $(w,h)$ represents the width and height of the detected box, and $(c_{x}, c_{y})$ is the top-left corner coordinate of the box.

\emph{3D pose and depth estimation.} 
A common approach to 3D pose estimation is regression, \eg, to estimate the quaternion~\cite{kendall2017geometric} or the elements of rotation matrices~\cite{Do2017deep6dpose}. However, in the preliminary experiments, we observed that when the object undergoes severe occlusion, generating a single reliable pose estimate is challenging: Inspected at only small portions of objects, indeed, significantly different pose configurations can be instantiated from visually very similar depth values (see Fig.~\ref{fig:distribution}(b) for example). To circumvent this challenge, we first generate multiple pose hypotheses and disambiguate them afterwards by exploiting the context of neighboring detections and their pose hypotheses during joint registration. To facilitate hypothesis generation, we adopt the classification approach of Kehl~\etal~\cite{Kehl_2017_ICCV}: Three rotation components are estimated separately, each represented as a classification problem with different classes binning an angle range: We use 30, 13, and 30 classes for pitch, yaw, and roll, respectively. For objects with axial symmetry, the roll component is not estimated. Similarly, the 3D depth values are estimated via classification (into 140 bins). The standard soft-max cross-entropy loss is used for training ($L_P$ for 3D pose and $L_D$ for depth):
\vspace{-0.8em}
\begin{align}
\label{eqn:cross-entropy_loss}
\begin{split}
L_{P (D)}=\frac{1}{N}\sum_{i=1}^{N}\mathbf{y}_{i}^{*}\cdot \log(\mathbf{y}_{i}),
\end{split}
\end{align}
where $\mathbf{y}^{*}$ is the ground-truth in the form of one-hot encoding, $\mathbf{y}_{i}$ is the class estimation after softmax activation, and $N$ is the batch size. Combining the hypotheses across all rotation and depth dimensions, we obtain 1.638M possible combinations (1.75K for axially symmetric objects) each provided with a confidence value as a product of individual classification confidences. These confidences are used by the joint registration module along with the spatial configurations of neighboring detections. For computational efficiency, before joint registration, we sample only five hypotheses per object by first removing many similar hypotheses using NMS, and then selecting the remaining hypotheses with the five highest confidences.

\vspace{-0.8em}
\paragraph{Joint registration.}

\begin{figure}[t]
\centering
\includegraphics[width=\linewidth]{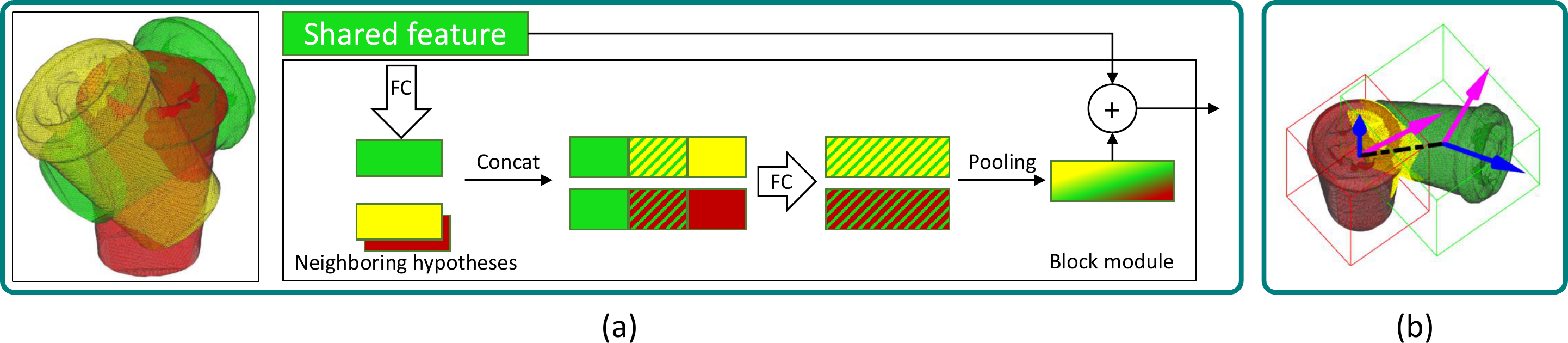}
\vspace{-0.3em}
\caption{\textbf{(a)} Schematic diagram of the feature extractor in the joint registration module (inspired by~\cite{Hosang_2017_CVPR}) applied to a hypothesis (green) with multiple neighboring hypotheses. Hatched boxes are the 6D relation features. Color in the box corresponds to hypothesis on the left image. Features from different hypothesis and 6D relation feature are concatenated. Variable number of neighbors for each hypothesis is reduced to a single feature by max-pooling. \textbf{(b)} An example neighborhood pattern: The intersection of the corresponding bounding boxes is highlighted in yellow; The black dotted line joints the center of bounding boxes; Blue and magenta arrows represent yaws and pitches, respectively.}
\label{fig:block_module}
\vspace{-1.0em}
\end{figure}

By design, the pool of hypotheses generated from 6D pose estimation module contains multiple false positives. As no prior information on the number of ground-truth object instances or their spatial configurations is available, the joint registration module needs to consider simultaneously the spatial configurations of multiple hypothesized object poses. We adapt GossipNet framework proposed by Hosang~\etal~\cite{Hosang_2017_CVPR} for this purpose: Applied to the bin-picking scenario, our GossipNet adaptation classifies each hypothesis of interest (HOI) into True positive ($1$) or False positive ($-1$) classes. The classification is made based on a \emph{contextual feature} encapsulating the spatial configurations of neighboring hypotheses (whose 3D bounding boxes overlap with the HOI).  Figure~\ref{fig:block_module}(a) visualizes this feature extraction process: First, for a given HOI (green), all neighboring hypotheses (NH; \eg red) are identified. Then, for each identified NH and HOI pair, an intermediate feature vector is generated by concatenating 1) the image depth features calculated within the bounding boxes of HOI and NH, respectively (solid green and red boxes in Fig.~\ref{fig:block_module}(a); See Fig.~\ref{fig:overview}(shared feature)) and 2) 13-dimensional feature vector of their joint configurations(6D relation feature): (1-2) confidence score of each detection, (3) intersection over union score of bounding boxes, (4-9) Euler angles normalized to $[0,1]$, 
(10-12) distances between the two detection centers along each $x$, $y$, and $z$ direction, normalized by the bounding box diameter, and (13) normalized $L_2$ distance between the two detection centers. 

The multiple intermediate pair-wise features (for all identified NHs) constructed in this way are then pooled to make a single contextual feature vector facilitating the classification independently of the number of neighboring hypotheses. During training, the registration network $f_J$ (combining the contextual feature extractor and a fully connected network classifier) minimizes the classification loss given below:
\vspace{-1.0em}
\begin{align}
\label{eqn:logistic_loss}
\begin{split}
L_J(f_J)=\sum_{i=1}^{N}a_{i}\cdot \log(1+\exp(-f(h_i)\cdot y_{i})).
\end{split}
\end{align}

Here, the batch-size $N$ is the number of total hypothesis $\{h_i\}$ in a single depth image, $y_{i}\in \{-1,1\}$ represents the ground-truths, 
and $a_{i}$ is a weight compensating the class imbalance decided at $1$ and $16$ for $1$ and $-1$ classes, respectively.

\begin{center}
\begin{figure}[t]
\centering
\includegraphics[width=\linewidth]{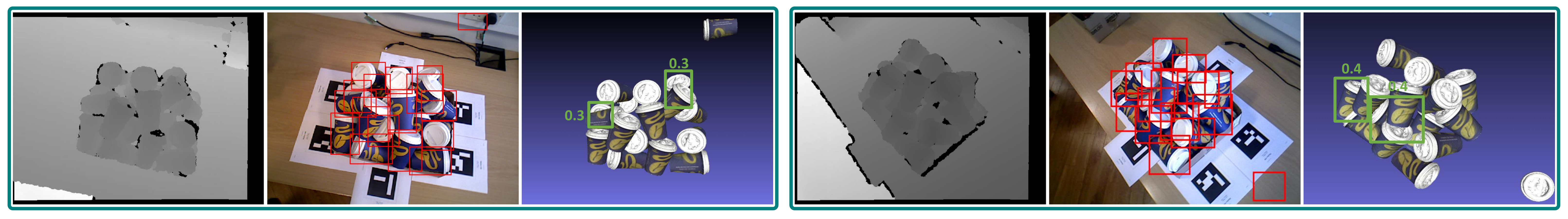}
\vspace{-0.5em}
\caption{Top $20$ hypotheses for the \emph{Coffee cup} scenario. For each triplet: (left) Input depth image. (middle) Detected bounding boxes. RGB image is used for visualization. (right) Reconstructed scene from hypotheses. Difficult objects are highlighted in green box annotated with visibility ratio. Objects with severe occlusion, some of which have visibility as low as $30 \%$, are successfully recovered.}
\label{fig:result_visualizations_coffee}
\vspace{-1em}
\end{figure}
\end{center}
\vspace{-2.7em}

\vspace{-1em}
\section{Experiments}
\label{sec:Experiments}
\vspace{-0.3em}
We evaluate our network on four objects from two different bin-picking datasets where multiple object instances occlude each other, reflecting realistic application cases: \textit{Coffee cup} from the Bin-Picking dataset~\cite{doumanoglou2016recovering}, and \textit{Bunny}, \textit{Pepper}, and \textit{Candlestick} from the Sil\'eane dataset \cite{bregier2017iccv}. We compare with existing algorithms:  Linemod algorithm~\cite{hinterstoisser2011linemod}; 3D point-pair-feature (PPF)-based algorithm~\cite{drost2010model}; and Latent-class Hough forest (LCHF)~\cite{tejani2014latent}. While PPF and LCHF are known to be robust to scene clutters and occlusions, respectively, they are not specifically designed for bin-picking applications. Therefore, we also compare with hypotheses verification approach~\cite{aldoma2012global} that extends Linemod and PPF for bin-picking applications (referred to as `Linemod+PP' and `PPF+PP', respectively). To gain an insight into the effectiveness of our multi-task learning approach in general 6D pose estimation (non bin-picking cases), we also evaluate our algorithm on the Domestic Environment dataset (\textit{Coffee cup} object)~\cite{tejani2014latent} providing examples of multiple object instances under mild occlusions. 

The accuracy of predictions is measured in average precision (\emph{AP}; equivalently, the area under the precision-recall curve) evaluated based on \emph{Sym} criteria adopted from~\cite{bregier2017iccv}: Based on \emph{Sym}, a hypothesis is accepted (considered as a true positive) if the distance to ground-truth is smaller than $0.1$ times the diameter of the smallest bounding sphere. We calculated the AP values of our algorithm using the evaluation code shared by the authors of~\cite{bregier2017iccv} enabling a direct comparison with the results of Linemod(+PP) and PPF(+PP) supplied from \cite{bregier2017iccv}. We also measure the AP based on \emph{ADD} criteria~\cite{hinterstoisser2012model} which accepts a hypothesis if its average distance from the ground-truth pose is less than $0.1$ times the object diameter. 

\begin{figure}[t]
\centering
\includegraphics[width=\linewidth]{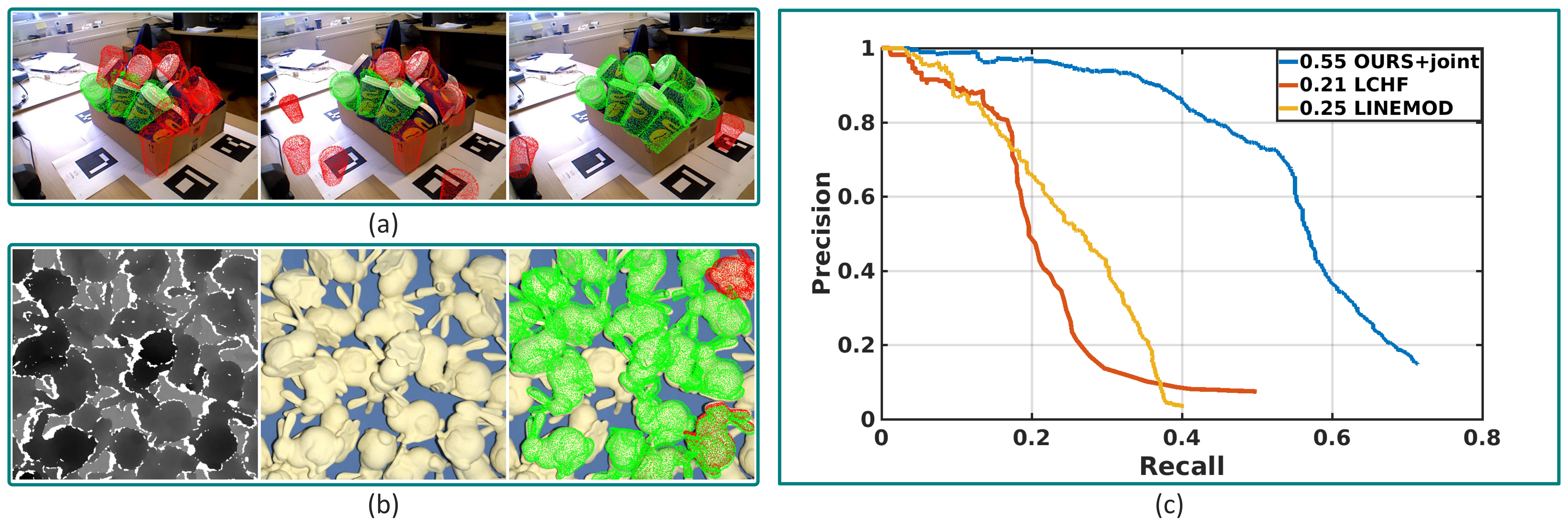}
\vspace{0.2em}
\caption{
\textbf{(a)} Example 6D pose estimates by Linemod, LCHF, and Ours, respectively (left to right). Correctly estimated poses are highlighted in green. 
\textbf{(b)} Top 20 hypotheses generated by our algorithm for a \emph{Bunny} image: (left) and (middle) The input depth image and the corresponding RGB image (for visualization); (right) Hypotheses projected onto the RGB image. True and false positives are highlighted in green and red, respectively. 
\textbf{(c)} The precision and recall curves of Linemod, LCHF, and our algorithm on \emph{Coffee cup} dataset. 
}
\label{fig:pr_curve}
\end{figure}
\vspace{-1em}
\begin{table*}[t]
\caption{\label{tab:result}\textbf{(a-b)} AP values of different algorithms: Each pose hypothesis is evaluated in \emph{Sym} criterion \textbf{(a)} and \emph{ADD} criterion \textbf{(b)}. F1 scores for Domestic Environment dataset (\emph{Cup} object)~\cite{tejani2014latent} with \textit{ADD}.}
\vspace{0.8em}
\centering
\setlength\tabcolsep{6pt}
\begin{minipage}{0.47\textwidth}
\centering

\resizebox{0.99\columnwidth}{!}{
\begin{tabular}{ l c c c c}
    \toprule
    \textbf{Method} & \emph{Bunny}~\cite{bregier2017iccv} & \emph{cup}~\cite{doumanoglou2016recovering} & \emph{pepper}~\cite{bregier2017iccv} & \emph{C.stick}~\cite{bregier2017iccv}\\ 
    \midrule
    Linemod \cite{hinterstoisser2011linemod}          & $0.39$ &$0.08$ &$0.04$ &$0.38$ \\ 
    Linemod+PP \cite{aldoma2012global}        & $0.45$ &$0.20$ &$0.03$ &$0.49$ \\
    PPF \cite{drost2010model}               & $0.29$ &$0.28$ &$0.06$ &$0.16$\\ 
    PPF+PP~\cite{aldoma2012global}            & $0.37$ &$0.30$ &$0.12$ &$0.22$\\
    \midrule
    Ours (w/o offset) & $0.49$ &$0.53$ &$0.32$ &$0.53$\\
    Ours       & $\mathbf{0.74}$ &$\mathbf{0.55}$ &$\mathbf{0.43}$ &$\mathbf{0.64}$\\
    \bottomrule
  \end{tabular}
}
\caption*{(a)}
\label{tab:result01} 
\end{minipage}%
\begin{minipage}{0.33\textwidth}
\centering
\resizebox{0.88\columnwidth}{!}{
  \begin{tabular}{ l c c }
    \toprule
    \textbf{Method} & Cup~\cite{doumanoglou2016recovering} & Cup~\cite{doumanoglou2016recovering}\\
                    & (real) & (synt.)\\
    \midrule
    Linemod~\cite{hinterstoisser2011linemod}         & $0.38$ & 0.32\\
    LCHF~\cite{tejani2014latent}            & $0.30$ & 0.31\\
    SSD6D~\cite{Kehl_2017_ICCV}            & $0.07$ & 0.01\\
    \midrule
    Ours            & $\mathbf{0.68}$ & $\mathbf{0.80}$ \\
    \bottomrule
  \end{tabular}
}
\caption*{(b)}
\end{minipage}%
\begin{minipage}{0.21\textwidth}
\centering

\resizebox{0.99\columnwidth}{!}{
    \begin{tabular}{ l c }
    \toprule
    \textbf{Method} & Cup~\cite{tejani2014latent} \\ 
    \midrule    
    Linemod~\cite{hinterstoisser2011linemod}                      & $0.819$ \\
    PPF~\cite{drost2010model}                 & $0.867$ \\
    LCHF~\cite{tejani2014latent} & $0.877$ \\
    Sahin~\etal~\cite{Sahin2017Learning}                 & $0.793$\\
    Kehl~\etal~\cite{kehl2016deep}                  & $0.972$ \\
    \midrule
    Ours                         & $\mathbf{0.992}$ \\
    \bottomrule

  \end{tabular}
  }
\caption*{(c)}
\end{minipage}%
\end{table*}
\vspace{-0.9em}

\vspace{1em}
\paragraph{Results on bin-picking datasets.}\hspace{-0.3cm}\footnote{As the original Bin-Picking dataset contained erroneous ground-truth annotation, every frame has been manually re-annotated via visual inspection.}
Table~\ref{tab:result}(a) summarizes the AP scores. No clear winner was found among the two classical algorithms, Linemod and PPF: Each one is better than the other on two datasets. As these algorithms are not designed for heavily cluttered-scenes, applying extra verification~\cite{aldoma2012global} significantly improved performance (Linemod+PP and PPF+PP). None of these four existing methods accounts for the occlusion patterns during training. Instead, they use the ICP to verify and refine the estimated hypothesis. By explicitly and simultaneously learning the occlusion patterns and object shapes during training, our algorithm (Ours) constantly outperformed existing algorithm with a large margin (15-31\%) even without having to rely on the external ICP algorithm.\\ 
\indent For qualitative comparison, Fig.~\ref{fig:pr_curve}(a) shows example detections: LCHF, Linemod, and Ours all correctly estimated the 6D poses of unoccluded foreground objects. However, LCHF and Linemod missed multiple objects with only mild levels of occlusions (highlighted in red), while our algorithm correctly recovered all objects with only two false positives leading to constantly and significantly higher precision across different recall values (Fig.~\ref{fig:pr_curve}(b)).

Due to severe occlusions caused by multiple instances of a single object, estimating the 6D poses in bin-picking using pure RGB-based approaches is challenging. We verified this by applying the state-of-the-art SSD6D algorithm~\cite{Kehl_2017_ICCV} in this scenario\footnote{We used the code kindly shared by the authors of this work.} (Table~\ref{tab:result}): Not surprisingly, the results of SSD6D algorithm failed to match the level of any depth image-based pose estimation algorithms in comparison.

`Ours (w/o offset)' in Table~\ref{tab:result}(a) shows the AP score of our algorithm obtained without the object center estimation to assess its contribution: Under severe occlusion, the initial bounding box detections are not tightly aligned with the ground-truths. Therefore, excluding this component severely degraded the joint registration process.

We also evaluate the performance of our algorithm with \textit{ADD} (Table~\ref{tab:result}(b)), which reconfirms the benefit of our explicit occlusion pattern learning approach. Note that both Linemod and LCHF, in comparison, use RGBD data whereas ours uses only depth images.
\vspace{-0.5em}
\paragraph{Non bin-picking scenario.} We use the F1 score following publicized results supplied in \cite{kehl2016deep,Tejani2018latent} (Table~\ref{tab:result}(c)): Even though our algorithm was trained for bin-packing scenes (with training data consisting of only foreground scenes), and it does not use the ICP refinement in contrast to others, it significantly outperformed Linemod~\cite{hinterstoisser2011linemod}, PPF~\cite{drost2010model}, and LCHF~\cite{Sahin2017Learning} and is on par with the state-of-the-art kehl~\etal's algorithm~\cite{kehl2016deep} demonstrating the effectiveness and robustness (to unseen background clutters) of our algorithm.

\vspace{-0.9em}
\paragraph{Implementation details and complexities.} Our shared feature extractor (Fig.~\ref{fig:overview}) is initialized with Imagenet-trained VGG network~\cite{russakovsky2015imagenet}. The entire network was trained using the ADAM optimizer with $10^{-5}$ learning rate. Training the network on each object took around 10 hours in a machine with a GTX1080ti GPU. Testing a single depth image took $204$ms.

\vspace{-1em}
\section{Conclusion}
\vspace{-0.7em}
Object detection and pose estimation in bin-packing scenarios is challenging due to severe occlusions caused by multiple instances of the same object. Existing approaches are limited in that their estimators are trained on isolated object instances and they do not directly consider possible mutual occlusion patterns during training. We presented a new 6D pose estimation framework that is trained directly on the scenes demonstrating multiple occlusion patterns. Our framework is instantiated as 2D detection, 6D pose hypothesis generation, and joint registration submodules which are jointly trained via a new multi-task losses. The resulting algorithm significantly outperformed the state-of-the-art 6D object pose estimation approaches on challenging datasets.

\paragraph{Acknowledgement:} This work is part of Imperial College London-Samsung Research project, supported by Samsung Electronics.

\bibliography{egbib}
\end{document}